\documentclass[final,nonatbib]{article}

\usepackage[final]{timeseries_workshop}

\usepackage[utf8]{inputenc} 
\usepackage[T1]{fontenc}    
\usepackage{hyperref}       
\usepackage{url}            
\usepackage{booktabs}       
\usepackage{amsfonts}       
\usepackage{nicefrac}       
\usepackage{microtype}      
\usepackage{xcolor}         
\usepackage{amsmath} 
\usepackage{ulem} 
\usepackage{tabularx}
\usepackage{graphicx}
\useunder{\uline}{\ul}{}
\title{Text2Freq: Learning Series Patterns from Text via Frequency Domain}

\author{%
  Ming-Chih Lo, Ching Chang, Wen-Chih Peng\\
  Department of Computer Science\\
  National Yang Ming Chiao Tung University\\ 
}

\begin{document}

\maketitle

\begin{abstract}
Traditional time series forecasting models mainly rely on historical numeric values to predict future outcomes.
While these models have shown promising results, they often overlook the rich information available in other modalities, such as textual descriptions of special events, which can provide crucial insights into future dynamics.
However, research that jointly incorporates text in time series forecasting remains relatively underexplored compared to other cross-modality work. 
Additionally, the modality gap between time series data and textual information poses a challenge for multimodal learning.
To address this task, we propose Text2Freq, a cross-modality model that integrates text and time series data via the frequency domain.
Specifically, our approach aligns textual information to the low-frequency components of time series data, establishing more effective and interpretable alignments between these two modalities.
Our experiments on paired datasets of real-world stock prices and synthetic texts show that Text2Freq achieves state-of-the-art performance, with its adaptable architecture encouraging future research in this field.

\end{abstract}

\section{Introduction}
\label{sec:intro}

The importance of incorporating textual information into time series forecasting is increasingly evident.
Real-world time series data is often influenced by external factors, such as news events, consumer feedback, and special occasions, which traditional models fail to account for \cite{DBLP:journals/corr/abs-2405-13522}.

\begin{figure}
  \centering
  \includegraphics[width=\linewidth]{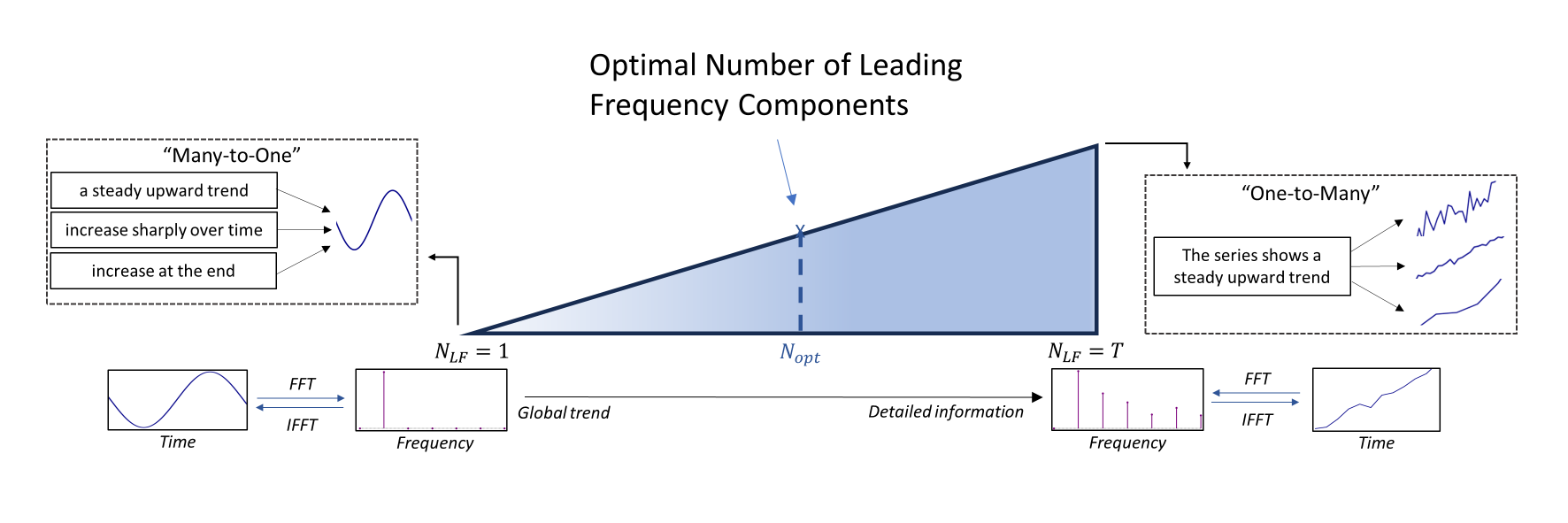}
  \caption{Illustration of alignment issues between time series and text. Starting with only the lowest frequency \(N_{LF} = 1\) captures slow-changing patterns like sinusoidal waves but loses details, causing many-to-one mapping issues from text to series. Increasing the frequency components to \(N_{LF} = T\) adds detail from a series but introduces noise, leading to one-to-many mapping issues. Since text encapsulates high-level patterns, this work aims to map text to an optimal subset of frequency components \(N_{opt}\) that balances noise and information loss.
}
  \label{fig:mapping relationship}
\end{figure}

Several approaches have been proposed in response to the growing need for multimodal learning in time series forecasting. 
However, these methods face three significant challenges.
First, the scarcity of paired datasets that combine time series and text makes the learning process difficult. 
Second, techniques from other cross-modality tasks, such as cross-attention, are hard to directly apply due to the significant modality gap between time series and text. 
This gap is due to differences such as text being discrete and rich in semantic content, while time series are continuous, focus on temporal changes, and often contain noise. 
Finally, text often encapsulates high-level patterns, such as overall trends, leading to a one-to-many problem when directly mapping text to time series (see Figure \ref{fig:mapping relationship}).

To overcome these challenges, we propose Text2Freq, a framework that align textual information to time series data through the frequency domain. 
Our approach includes a pre-trained text-to-frequency module, trained on a boarder dataset to address the issue of limited paired data. 
Additionally, by aligning text with the low-frequency components of time series, Text2Freq effectively bridges the modality gap and extracts clear patterns from text, enhancing interpretability.

We validate Text2Freq using real-world stock price data paired with synthetic text generated by GPT-4 \cite{DBLP:journals/corr/abs-2303-08774}.
To the best of our knowledge, Text2Freq is the first method to align textual information with low-frequency series components, leading to improved performance in mulitimodal forecasting with more effective alignment.

\section{Problem Formulation and Methodology}

\subsection{Multimodal Time Series Forecasting Enhanced by Textual Data}
Given a multivariate time series $X_{\text{past}} \in \mathbb{R}^{L \times C}$ consisting of $L$ past steps and $C$ channels, along with corresponding textual information \textit{Text} related to the target variable we aim to predict, our objective is to predict the future univariate series $X_{\text{future}} \in \mathbb{R}^{T}$.

\subsection{Model Architecture}
The proposed TextToFreq model consists of two stages: a text-to-frequency pre-trained model and the overall fusion model.

\subsubsection{Stage 1: The Pre-trained Text-To-Frequency Transformer}
\begin{figure}
  \centering
  \includegraphics[width=\linewidth]{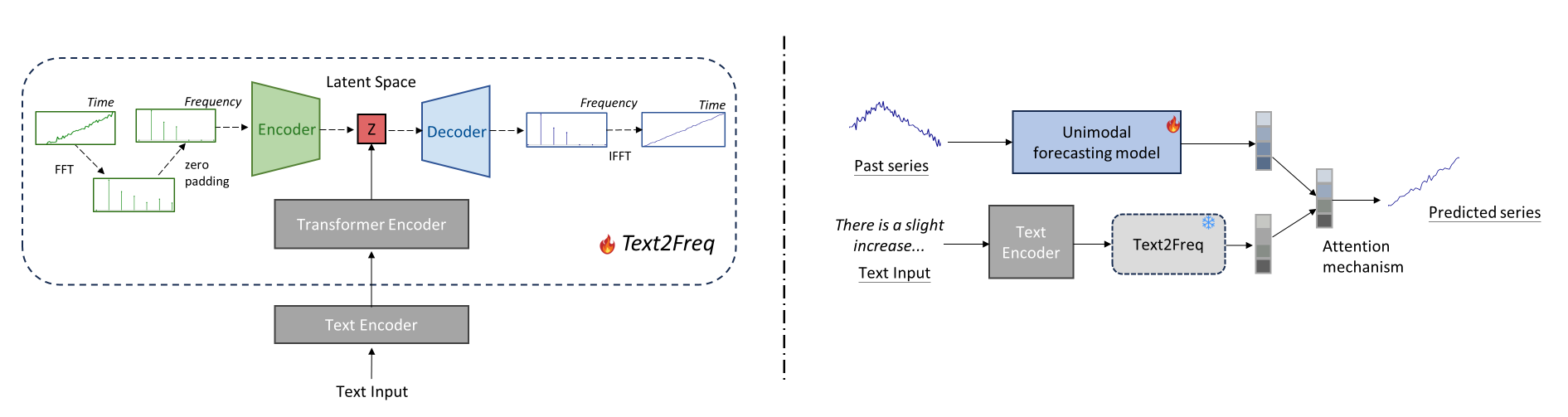}
  \caption{Overview of Text2Freq. Left panel: Stage 1 - Pre-training. Text embeddings are mapped to the latent space of frequency components using a Transformer Encoder. Right panel: Stage 2 - Multimodal Fusion. The pre-trained Transformer is frozen, and the outputs from both modalities are fused using an attention mechanism.
}
  \label{fig:model framework}
\end{figure}

As depicted in Figure \ref{fig:model framework}, we employ a pre-trained BERT \cite{DBLP:conf/naacl/DevlinCLT19} to extract text features. 
These embeddings are then mapped to the latent space of low-frequency components in a time series using a Transformer Encoder architecture \cite{DBLP:conf/nips/VaswaniSPUJGKP17}. 
The mapped embeddings are then decoded back into the time series domain for multimodal fusion in the next stage.
Specifically, we apply the Discrete Fourier Transform (DFT) to convert the time series into the frequency domain and extract the leading low-frequency components, excluding the DC component due to series normalization.
To enhance the latent space representation in the frequency domain, we utilize a Variational Autoencoder (VAE) \cite{DBLP:journals/corr/KingmaW13} architecture, as inspired by \cite{DBLP:conf/aistats/LeeMA23}.

\subsubsection{Stage 2: The Multimodal Fusion}
In this stage, we freeze the pre-trained Transformer Encoder from Stage 1 and focus on training the remaining components of our model: the unimodal time series forecasting model and the fusion layer.
Our objective is to effectively integrate information from both the text and the time series predictions into a unified representation, i.e., the time series.
Specifically, we employ patching and channel-independence \cite{DBLP:conf/iclr/NieNSK23} architecture for the unimodal forecasting model. 
The fusion layer then concatenates the output series from text-to-frequency pre-trained model with the output from the time series forecasting model.
This combined information is subsequently processed through an attention mechanism to generate the final prediction.

\section{Experiment: Stock Price Prediction}

\subsection{Dataset}
For our experiment, we utilized a weekly stock price dataset featuring Apple (AAPL), Microsoft (MSFT), and Google (GOOG) stocks from December 1, 2006, to November 30, 2016. 
To address the challenge of obtaining insightful textual information for each stock, we used GPT-4 to generate accurate descriptions of future patterns by providing it with the actual future series for each instance.

For our pre-trained model that aim to solve the limited dataset problem, we used the TRUCE dataset \cite{DBLP:journals/corr/abs-2110-01839}, which contains a more extensive range of paired time-series and text data. 
This dataset includes weekly stock prices from seven companies, totaling 1900 instances, each with a sequence length of 12. 
To enhance the model's capacity, we use the same way to augment the paired text data with descriptions generated by GPT-4.

\subsection{Experiment Setup}
We evaluate Text2Freq against two baseline approaches:  unimodal time series forecasting models that use only time series data, and multimodal models that integrate both time series and text data. For unimodal forecasting, we use PatchTST \cite{DBLP:conf/iclr/NieNSK23}, a representative model in time series forecasting, as our baseline. For multimodal forecasting, we employ a attention-based methodology similar to \cite{DBLP:journals/corr/abs-2405-13522},  where series data is processed through a unimodal forecasting model and text data through a Transformer Encoder, with outputs combined using an attention mechanism.
All models are configured with a look-back window of 36 and a prediction length of 12.

Furthermore, we compared the alignment by mapping the text sequence directly to the original time series with mapping it to the frequency domain, to evaluate whether the frequency domain provides a more effective learning framework.

Additionally, to validate the alignment phenomenon discussed in Introduction~\ref{sec:intro} and Figure~\ref{fig:mapping relationship}, we examined how varying amounts of low-frequency information extracted from the original series affect the alignment between text embeddings and time-series embeddings in the latent space.

Table  \ref{tab:comparison} and \ref{tab:frequency_ablation} summarize the results of our experiments. 

\begin{table}[h!]
\centering
    \begin{tabular}{l|c|c|c|c|c|c}
        \hline
        Methods & \multicolumn{2}{c|}{Text2Freq} & \multicolumn{2}{c|}{Attention Fusion} & \multicolumn{2}{c}{PatchTST} \\ \hline
        Metric  & MSE & MAE & MSE & MAE & MSE & MAE \\ \cline{2-7}
        MSFT    & \textbf{0.734} & \textbf{0.644} & {\ul 0.890} & {\ul 0.715} & 0.980 & 0.780 \\
        AAPL    & \textbf{0.160} & \textbf{0.300} & {\ul 0.237} & {\ul 0.364} & 0.368 & 0.482 \\
        GOOG    & \textbf{1.056} & \textbf{0.745} & {\ul 1.143} & {\ul 0.789} & 1.318 & 0.851 \\ \hline
        Avg.    & \textbf{0.650} & \textbf{0.563} & {\ul 0.757} & {\ul 0.623} & 0.889 & 0.704 \\ \hline
    \end{tabular}
    \vspace{0.1cm}
\caption{Forecasting comparison. We set the lookback window size L as 36 and the prediction length as 12. The best values are in bold and second best are underlined. 'Attention Fusion' refers to the multimodal forecasting baseline we use.}
\label{tab:comparison}
\end{table}

\begin{table}[h!]
\centering
\begin{tabularx}{\textwidth}{>{\centering\arraybackslash}X|>{\centering\arraybackslash}X|>{\centering\arraybackslash}X|>{\centering\arraybackslash}X|>{\centering\arraybackslash}X|>{\centering\arraybackslash}X|>{\centering\arraybackslash}X|>{\centering\arraybackslash}X}
\hline
Alignment & Text-Series & Text-Freq 1 & Text-Freq 2 & Text-Freq 3 & Text-Freq 4 & Text-Freq 5 & Text-Freq 6 \\ \hline
MSE     & 0.897       & 0.855       & 0.853       & \textbf{0.764} & {\ul 0.788} & 0.841       & 0.840       \\
MAE     & 0.734       & 0.738       & 0.730       & {\ul 0.686} & \textbf{0.682} & 0.710       & 0.700       \\ \hline
\end{tabularx}
\vspace{0.1cm}
\caption{Comparison of different mapping strategies. The MSE and MAE metrics are calculated based on the loss between the output series and the original series. 'Text-Series' maps text embeddings directly to the time-series latent space, while 'Text-Freq' maps text to the number of leading low-frequency components. Since our original series sequence length is 12, the maximum number of frequency components is 6. The best values are in bold and second best are underlined.}
\label{tab:frequency_ablation}
\end{table}

\subsection{Key Observations}

\subsubsection{Modeling with Text Improves Forecasting Performance}
Incorporating text into forecasting models significantly enhances performance. Our Text2Freq and the attention-based multimodal model reduce mean squared error (MSE) by 26\% and 14\%, respectively, compared to the unimodal time series model PatchTST. By integrating text, these models more effectively consider implicit future patterns, addressing the limitations of unimodal models.

\subsubsection{Aligning Text to Low-Frequency Components of a Series is Beneficial}
Our method, Text2Freq, shows a 14\% improvement in MSE compared to an attention-based multimodal model.
This underscores our assertion that directly learning from both time series and text can lead to a one-to-many problem, where text may align with noisy information that should be disregarded.

Additionally, by analyzing the effect of various low-frequency components from the original series on alignment (see Table \ref{tab:frequency_ablation}) using the TRUCE dataset, we observe that including all frequency components during mapping can cause one-to-many mapping issues, as previously noted.
In contrast, using only the lowest frequency components can lead to a many-to-one problem, where multiple pieces of textual information map to similar wave patterns.
Both scenarios present challenges for convergence and can result in either missing or noisy information.
Therefore, by carefully selecting the optimal amount of low-frequency components, we can better capture the global trend patterns of a series, improving model convergence and addressing alignment issues.

\subsubsection{Learning in Frequency Domain Yields Better Alignment}
Learning series patterns from the frequency domain offers advantages like a global view and energy compaction \cite{DBLP:conf/nips/YiZFWWHALCN23}. 
This approach enhances the extraction of high-level patterns and key components from a series. 
As shown in Table \ref{tab:frequency_ablation}, mapping via the frequency domain (with all frequency components) surpasses direct text-to-series mapping by over 6\% in MSE, demonstrating that the frequency domain serves as a more effective medium for bridging the modality gap between text and time series.

\section{Conclusion and Future Works}
In this work, we introduce Text2Freq, a novel approach that integrates textual information with time series through a frequency-domain learning process.
Our evaluation using a real-world stock price dataset demonstrates that mapping text to the low-frequency components of the series and combining this with series predictions significantly enhances time series forecasting. 
Furthermore, we believe that integrating our pre-trained framework with various advanced models from the fields of time series forecasting and natural language processing, such as foundation models, could further improve the performance and interpretability of multimodal learning.
Future work will focus on effectively combining models across different modalities to optimize overall performance in diverse scenarios.

\small
\bibliographystyle{unsrt}
\bibliography{reference}

\medskip



\newpage

\appendix

\section{Related Works}

\paragraph{Cross-modality Learning in Time Series}
Recent advances in multimodal studies have explored integrating text with various data types, such as images \cite{DBLP:conf/icml/RadfordKHRGASAM21} and audio \cite{DBLP:journals/taslp/YangYWWWZY23}.
However, the intersection of text and time series remains relatively underexplored, This is largely due to the scarcity of paired datasets and the fundamental differences between time series data and textual information.
Current approaches mainly transform or reprogram time series into a text modality \cite{DBLP:journals/corr/abs-2308-08469}, \cite{DBLP:conf/iclr/0005WMCZSCLLPW24}, \cite{ansari2024chronos} to leverage large language models for forecasting or other downstream tasks. 
Despite these efforts, significant advancements in transforming text into time-series remains limited.

\paragraph{Integrating Text with Time Series Forecasting}
Research on text-guided time series forecasting is still emerging, with a few notable contributions. 
TEMPO \cite{DBLP:conf/iclr/CaoJAPZY024} integrates textual information by decomposing series and using a transformer architecture. 
TGForecaster \cite{DBLP:journals/corr/abs-2405-13522} employs PatchTST \cite{DBLP:conf/iclr/NieNSK23} as a backbone model with a cross-attention mechanism to incorporate text into forecasting.
Text2TimeSeries \cite{DBLP:journals/corr/abs-2407-03689} integrates real-world events into time series predictions, refining stock price forecasts by mapping event-induced changes to directional price movements.
While these approaches improve performance over unimodal models that only take time series as inputs, they do not fully address the modality gap or the text-series mapping relationship.
In this work, we bridge the divergence between time series and text by proposing a framework that integrates textual information into time series forecasting in a more interpretable and effective manner.

\section{Data Generation: Synthetic Textual Information Generated by GPT-4
}
As discussed in the previous section, there is a shortage of insightful and sufficiently paired text and time series data. To address this, we utilize GPT-4 to generate ground-truth pattern descriptions for each target future series. The structure of the prompts used for generation is as follows:

\begin{figure}[h]
  \centering
  \includegraphics[width=\linewidth]{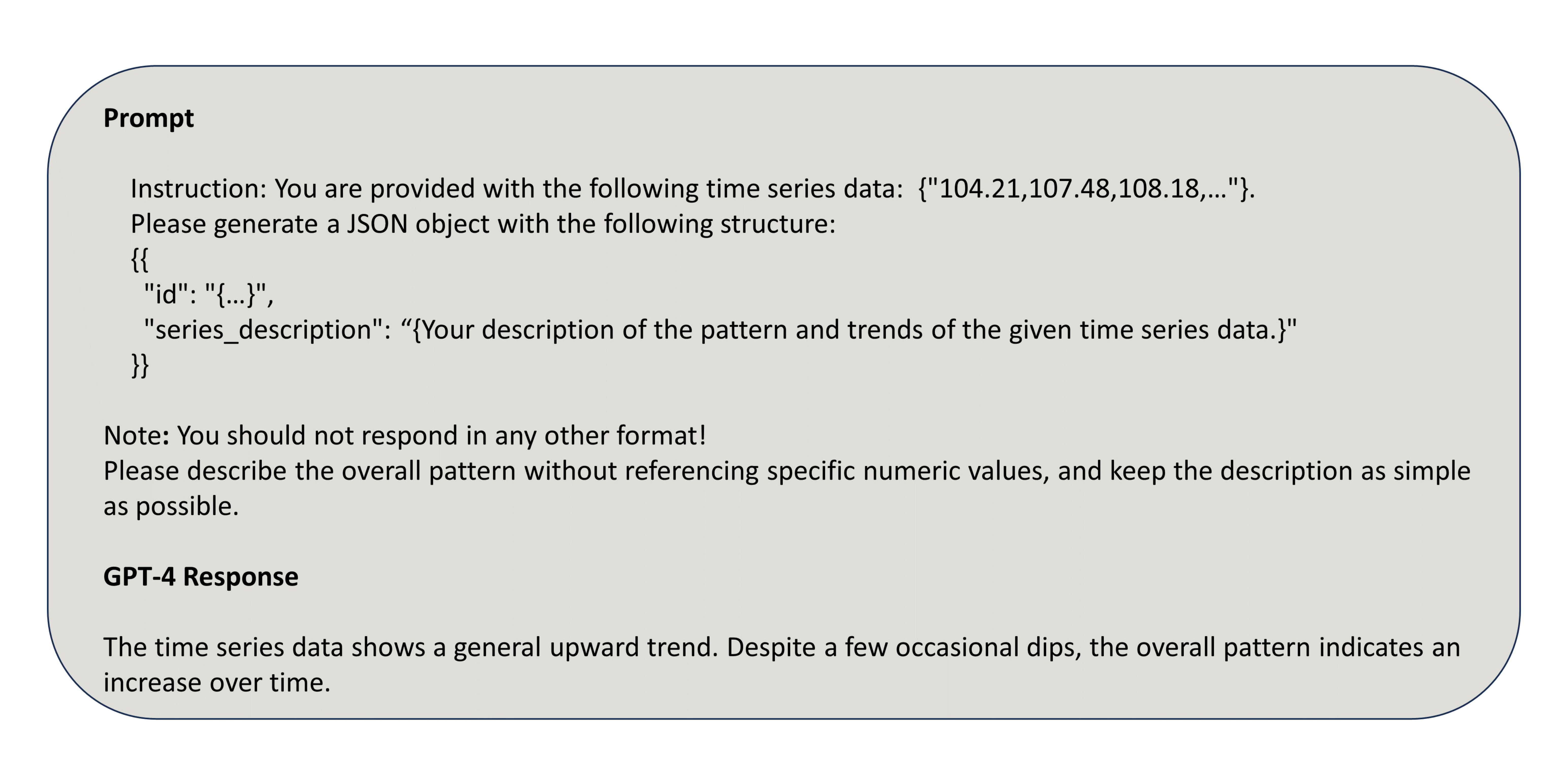}
  \caption{The prompt structure for data generation. 
}
  \label{fig:Prompt}
\end{figure}

Future work will also concentrate on effectively extracting textual information from a variety of sources across different datasets to further validate the capabilities of our methods.

\newpage
\section{Detail Explanation of Model Architecture}

As outlined in the main section, the proposed Text2Freq model comprises two stages: the pretraining phase and the overall fusion model. Below, we provide further details on the architecture of each component.

\paragraph{Details of the Pretrained Text-to-Frequency Transformer} \mbox{}\\\mbox{}
As shown in Figure \ref{fig:model framework}, Stage 1 of the pretraining process includes two primary substeps: constructing a Variational Autoencoder (VAE) architecture to learn the latent space of time series data and using a Transformer Encoder to align text embeddings with corresponding series latent representations.

Inspired by \cite{DBLP:conf/aistats/LeeMA23}, we design a VAE to capture the latent space of frequency components within time series data. First, we transform each series to its frequency spectrum using the Discrete Fourier Transform (DFT). To focus on low-frequency information, we apply zero-padding to the higher frequencies, preserving a predefined number of low-frequency components. These components are then fed into the VAE, where the encoder compresses the input to a latent representation, and the decoder reconstructs the frequency spectrum from this latent space.

To align text with series data, we adapt the approach in [2]. Text features are extracted from a pre-trained BERT model, providing robust embeddings that capture semantic nuances. These embeddings are then mapped to the latent space of the leading low-frequency components of the series using a Transformer Encoder, effectively aligning the text and series representations.

\paragraph{Details of the Overall Multimodal Fusion}\mbox{}\\\mbox{}
Following the first stage of pretraining, we freeze the pretrained Text2Freq model and assess its effectiveness within a multimodal framework that integrates both text and time series inputs to evaluate combined performance.

For the time series input, we process the past series data through a unimodal forecasting model, as shown on the right side of Figure \ref{fig:model framework}. This model is designed with channel independence and a patching structure \cite{DBLP:conf/iclr/NieNSK23} to handle sequential data effectively.

For the text input, we begin by extracting features using a pre-trained BERT model, which captures the semantic characteristics of the text. These text embeddings are then passed through the frozen Text2Freq model to generate corresponding series patterns.

Once we obtain outputs from both modalities, we fuse them using an attention mechanism to allow effective interaction between the time series and text representations, enhancing multimodal predictive performance.

\section{Limitation}
We validate the effectiveness of the Text2Freq model using stock price data. Experimental results indicate that aligning text embeddings to the frequency domain significantly outperforms existing methods. However, we acknowledge a limitation in the dataset, as it contains information leakage: the text input used in stage 2 is based on the ground truth description of future patterns, due to limited available data. To address this, future work will focus on evaluating our model with real-world textual data sources, such as news articles or event descriptions, to ensure robust validation.

In this paper, we also highlight the inherent challenges of aligning time series data with textual information. This alignment task presents opportunities for further research on multimodal fusion in time series forecasting, advancing the integration of diverse data types for improved forecasting accuracy.

\end{document}